\documentclass[letterpaper]{article} 
\usepackage{aaai25}  
\usepackage{times}  
\usepackage{helvet}  
\usepackage{courier}  
\usepackage[hyphens]{url}  
\usepackage{graphicx} 
\usepackage{natbib}  
\usepackage{caption} 
\usepackage{algorithm}
\usepackage{algorithmic}

\usepackage{newfloat}
\usepackage{listings}
\DeclareCaptionStyle{ruled}{labelfont=normalfont,labelsep=colon,strut=off} 
\floatstyle{ruled}
\newfloat{listing}{tb}{lst}{}
\floatname{listing}{Listing}
\usepackage{amsmath}
\usepackage{amssymb}

\title{Adaptive Sparsified Graph Learning Framework for Vessel Behavior Anomalies}
\author{
    Jeehong Kim\textsuperscript{\rm 1}\equalcontrib, Minchan Kim\textsuperscript{\rm 1}\equalcontrib, Jaeseong Ju\textsuperscript{\rm 1}, Youngseok Hwang\textsuperscript{\rm 1},Wonhee Lee\textsuperscript{\rm 2}, Hyunwoo Park\textsuperscript{\rm 1}\thanks{Corresponding author}
}
\affiliations{
    \textsuperscript{\rm 1}Graduate School of Data Science, Seoul National University \\
    \textsuperscript{\rm 2}Korea Research Institute of Ships and Ocean Engineering\\

    \{williamkim10, mmm5373, hyunwoopark\}@snu.ac.kr
}

\begin{document}

\maketitle

\begin{abstract}
Graph neural networks have emerged as a powerful tool for learning spatiotemporal interactions. However, conventional approaches often rely on predefined graphs, which may obscure the precise relationships being modeled. Additionally, existing methods typically define nodes based on fixed spatial locations, a strategy that is ill-suited for dynamic environments like maritime environments. Our method introduces an innovative graph representation where timestamps are modeled as distinct nodes, allowing temporal dependencies to be explicitly captured through graph edges. This setup is extended to construct a multi-ship graph that effectively captures spatial interactions while preserving graph sparsity. The graph is processed using Graph Convolutional Network layers to capture spatiotemporal patterns, with a forecasting layer for feature prediction and a Variational Graph Autoencoder for reconstruction, enabling robust anomaly detection.
\end{abstract}


\section{Introduction}
Graph neural network (GNN) based methods provided a powerful approach for modeling spatiotemporal data \cite{yu2017spatio}, yet existing approaches often relied on predefined or fully connected graphs \cite{liu2023model, zhang2023vessel}, introducing noise and reducing interpretability. Additionally, most graph based anomaly detection methods primarily utilized graphs for node embedding updates, rather than fully utilizing their structural and relational properties \cite{liu2023model, zhang2023vessel}.

We introduce a novel framework for anomaly detection in multi-ship trajectory data by utilizing a sparsified graph, removing noisy and task-irrelevant edges while retaining essential spatiotemporal relationships. The optimized graph is then embedded using Graph Convolutional Network (GCN) \cite{kipf2016semi} layers to effectively capture these relationships. Furthermore, leveraging a Variational Graph Autoencoder (VGAE) \cite{kipf2016variational}, our framework reconstructs adjacency matrices, allowing anomaly detection by identifying discrepancies in graph structures.

Our contributions are summarized as follows:
\begin{itemize}
    \item \textit{Novel Graph Representation}: We establish a graph-based input structure for moving objects, representing each timestamp as a node to explicitly model temporal dependencies through graph edges.
    \item \textit{Efficient Graph Construction Strategy}: We generate a multi-ship graph that effectively captures spatial relationships while preserving graph sparsity.
    \item \textit{Graph Based Anomaly Detection}: Our framework combines prediction and reconstruction within a graph based approach, fully leveraging graph structures to enhance anomaly detection performance.
\end{itemize}

\begin{figure*}[t]
\centering
\includegraphics[width=0.95\textwidth]{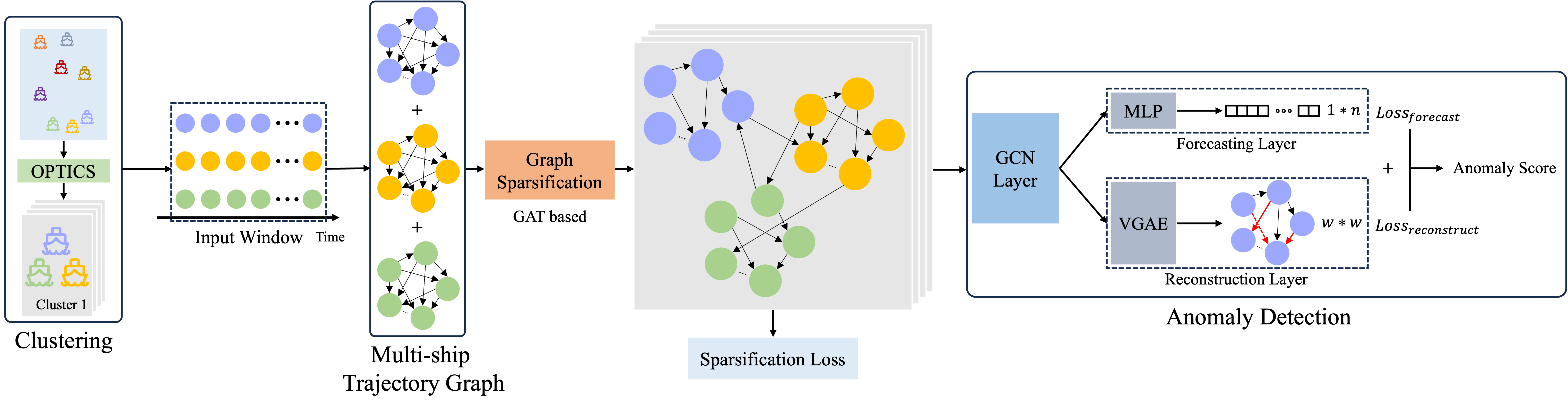}
\caption{ 
Our framework employs a multi-component architecture designed to learn a sparsified graph structure. The learned graph is then embedded using GCN layers, which supports both the forecasting and reconstruction processes. The combined loss from these components is utilized for effective anomaly detection.
}
\label{fig2}
\end{figure*}
\section{Related Works}
\paragraph{Spatiotemporal GNNs}
Existing spatiotemporal methods often relied on fixed spatial locations, such as intersections \cite{wang2020traffic, yu2017spatio}, which were unsuitable for the dynamic and fluid nature of maritime environments. Consequently, although efforts were made to establish dynamic reference points that were appropriate for the maritime domain \cite{eljabu2021anomaly, liang2022fine}, these points still did not adequately capture the fluid and constantly evolving nature of moving vessels.

\paragraph{Vessel Behavior Anomaly}
The concept of anomalies in AIS tracks refers to behaviors that deviate from what is considered ‘normal’ or expected under typical operational conditions \cite{laxhammar2008anomaly}. There are many studies \cite{lane2010maritime, davenport2008kinematic, liu2024ais} that define anomalous behaviors based on kinematic behaviors,  AIS transmission behaviors, and other supplementary behaviors that occur on the ship. One of the key challenges in defining anomalies is the absence of a universal criterion for what constitutes an anomalous event. Despite the wide range of possible anomalies, this study specifically focuses on \textit{deviation from standard route}. This type of anomaly is one of the most fundamental and frequently observed irregularities in vessel movement, serving as a crucial indicator of potential maritime risks. A vessel straying from its expected trajectory could signal various underlying causes, including adverse weather conditions, mechanical failures, unauthorized maneuvers, or illicit activities. By analyzing deviations from standard routes, we aim to establish a robust framework for detecting navigational anomalies in real-world maritime operations. The rationale behind this selection will be elaborated more in the experimental section, where we detail the statistical techniques used to quantify deviations.

\paragraph{GNN Based Vessel Anomaly Detection}
The objective of vessel anomaly detection is to identify unusual movement patterns, which are often caused by mechanical failures or navigational errors \cite{ribeiro2023ais}. Most existing methods constructed graphs using predefined or fully connected structures \cite{jiang2024stmgf, liu2023model, wolsing2022anomaly, zhang2023vessel}. Nevertheless, these methodologies proved inadequate for capturing meaningful spatiotemporal relationships that are well-suited to the task at hand.

\section{Methodology}
\subsection{Problem Definition}
Given AIS data $\mathcal{X} = \{\mathbf{x}_i^t | i \in \mathcal{V}_t, t \in T\}$, where $\mathbf{x}_i^t$ represents the feature vector of ship $i$ at time $t$, and a series of temporal graphs $\{\mathcal{G}_t\}_{t=1}^T$, the objective is to:
\begin{enumerate}
    \item Predict the \textit{next time step embedding} $\mathbf{h}_i^{t+1}$ for each node $i \in \mathcal{V}_t$ using past observations and graph structure.
    \item Identify anomalous nodes and edges in $\mathcal{G}_{w_t}$ by comparing observed patterns to learned normal behavior.
\end{enumerate}

The proposed approach combines \textit{node embedding}, \textit{graph sparsification}, \textit{time-series forecasting}, and \textit{anomaly detection} into a unified framework to effectively handle these challenges.

As depicted in Figure 1, our paper establishes a novel framework for vessel anomaly detection leveraging graph neural networks to model multi-ship trajectory data.

\subsection{Graph Construction}
To define the boundaries of multi-ship graphs, the OPTICS clustering algorithm \cite{ankerst1999optics} is employed. This algorithm groups ships based on the latitude and longitude at a given time stamp $t$, thereby identifying clusters that reflect shared spatiotemporal regions. 

\paragraph{Graph Initialization}
Unlike conventional GNN tasks where the initial graph structure is predefined, our approach starts by constructing the graph from raw data. In our case, the graph is constructed within each identified cluster from the previous step. The core idea is illustrated on lines 4–5 of Algorithm~\ref{alg:Graph Initialization}, where each edge represents a temporal connection between consecutive timestamps.

The underlying intuition behind this approach is to demonstrate that, rather than relying on traditional time-series prediction models such as RNN \cite{rumelhart1986learning} or LSTM \cite{hochreiter1997long}, temporal dependencies can be effectively captured within a graph structure. By embedding sequential relationships as edges in the graph, we establish a framework where time-series forecasting can be performed using a simple MLP model. This highlights the feasibility of leveraging graph-based representations for temporal modeling, enabling efficient and scalable predictions without the need for complex recurrent architectures.
\begin{algorithm}[tb]
\caption{Graph Initialization}
\label{alg:Graph Initialization}
\textbf{Input}: Timestamps of ship $i$: $t_1^i, t_2^i, \dots, t_n^i$, window size $w$ \\
\textbf{Output}: Graphs $\mathcal{G}_{w_1}^i, \mathcal{G}_{w_2}^i, \dots, \mathcal{G}_{w_m}^i$ 

\begin{algorithmic}[1]
\FOR{each sliding window $w_k$}
    \STATE Initialize graph $\mathcal{G}_{w_k}^i = (\mathcal{V}_{w_k}^i, \mathcal{E}_{w_k}^i)$
    \STATE $\mathcal{V}_{w_k}^i \gets$ timestamps in $w_k$
    \FOR{each pair of timestamps $(t_1, t_2)$ where $t_1, t_2 \in w_k$ and $t_1 < t_2$}
        \STATE Add directed edge $e_{t_1 \to t_2}$ to $\mathcal{E}_{w_k}^i$
    \ENDFOR
\ENDFOR
\RETURN $\mathcal{G}_{w_1}^i, \mathcal{G}_{w_2}^i, \dots, \mathcal{G}_{w_m}^i$
\end{algorithmic}
\end{algorithm}

\paragraph{Multi-ship Trajectory Graph}
Then the multi-ship trajectory graph is constructed in the same context for a single ship while only considering the ships within the same cluster. This guarantees that the graph encompasses both intra-ship and inter-ship temporal interactions.

\subsection{Graph Sparsification}
Graph sparsification is essential for efficient computation, particularly when the graph is initially constructed using Algorithm~\ref{alg:Graph Initialization}. Algorithm~\ref{alg:Graph Sparsification} focuses on learning an edge-sparsified graph by systematically removing noisy and task-irrelevant edges. It is based on SGAT \cite{ye2021sparse}, which employs a binary mask $Z \in \{0, 1\}^M$, where $ M$ denotes the total number of edges. The function of the mask $z_{ij}$ is to determine whether the edge $e_{ij}$ is used during neighbor aggregation. The adjacency matrix $A$ is modified as $\bar{A} = A \odot Z$ and sparsification is achieved by optimizing a regularized loss of the $L_0$ norm.

While there is a separate regularization for this process, recent approaches \cite{li2024gslb} suggest integrating the term directly with the downstream task loss, which could offer a more unified optimization framework. Further advancements in this direction can be explored through the emerging field of \textit{Graph Structural Learning}, which has gained significant attention in recent research. 

\begin{algorithm}[tb]
\caption{Graph Sparsification}
\label{alg:Graph Sparsification}
\textbf{Input}: Graph $\mathcal{G}^i = (\mathcal{V}^i, \mathcal{E}^i)$ \\
\textbf{Output}: Sparsified Graph $\mathcal{G}_{\text{sparse}}^i$

\begin{algorithmic}[1]
\STATE Initialize sparsification function $F$
\STATE Apply $F$ to $\mathcal{G}_i$: 
\[
\mathcal{G}_i^{\text{sparse}} = F(\mathcal{G}_i)
\]
\STATE Perform downstream tasks using $\mathcal{G}_{\text{sparse}}^i$
\STATE Compute $L_{\text{forecast}}$ and $L_{\text{reconstruct}}$
\STATE Apply sparsity regularization with $L_0$ norm:
\[
L_{\text{total}} = L_{\text{forecast}} + L_{\text{reconstruct}} + \lambda \|Z\|_0
\]
\STATE Optimize $\mathcal{G}_{\text{sparse}}^i$ by minimizing $L_{\text{total}}$
\STATE \textbf{return} $\mathcal{G}_{\text{sparse}}^i$
\end{algorithmic}
\end{algorithm}

\subsection{Anomaly Detection}
The anomaly detection process begins with the sparsified graph derived from the multi-ship trajectory graph, which is then subjected to a series of GCN layers. These layers capture structural and spatiotemporal patterns, embedding them into node representations.

The output of the GCN layers is proceeded to two additional layers, namely the \textit{forecasting layer} and the \textit{reconstruction layer}. These stages are performed at the individual ship level for a tailored process, where each ship’s trajectory is treated as an independent subgraph.

The two-layer design integrates both forecasting and reconstructive perspectives, enabling comprehensive anomaly detection for temporal deviations and feature distortions. By leveraging the graph's inductive bias, specifically the directed edges that are constructed upon temporal dependencies, the architecture eliminates the need for sequential models like LSTM or GRU, ensuring a streamlined and robust design for diverse graph structures.

\begin{equation}
L_{\text{total}} = L_{\text{forecast}} + L_{\text{reconstruct}} + \lambda \|Z\|_0
\end{equation}

\paragraph{Forecasting} The forecasting layer employs a simple multi-layer perceptron (MLP) to predict the feature values at the subsequent time step for each node, utilizing the temporal dependencies encoded by the GCN layers. A discrepancy between the predicted and actual values may be indicative of a potential anomaly. 

\paragraph{Reconstruction} The reconstruction layer utilizes VGAE to process individual ship trajectories as distinct subgraphs, encoding their structural and feature information into a compact latent space. The VGAE reconstructs the adjacency matrix, with anomalies identified based on reconstruction errors derived from the differences between the original and reconstructed subgraph properties. By capturing subgraph level patterns, this approach facilitates robust anomaly detection for individual ship trajectories.

\begin{equation}
\begin{split}
L_{\text{reconstruct}} = \mathbb{E}_{q(Z|X, \bar A)}[\log p(\bar A, X|Z)] \\
- \beta \cdot D_{\text{KL}}(q(Z|X, \bar A) \| p(Z))
\end{split}
\end{equation}

\paragraph{Anomaly Scoring} To detect anomalous data points, we adopt a reasoning score that integrates both the prediction error and the node reconstruction probability \cite{liu2023model}. For each data point i, the reasoning score is defined as

\begin{equation}
    \text{RS}_i = \frac{E_i + \gamma \cdot \left(1 - P_i\right)}{1 + \gamma},
    \label{eq:reasoning_score}
\end{equation}

where $E_i$ denotes the mean squared error between predicted and actual values, $P_i$ is the node reconstruction probability representing the likelihood of observing the feature values under the learned model, and $\gamma$ is a hyperparameter that controls the balance between these two components. The optimal value of $\gamma$ can be determined through validation on the training set.

To determine the anomaly threshold, we employ the Peak-Over-Threshold (POT) method, which dynamically adjusts the threshold based on environmental changes. This approach classifies a data point as anomalous when its prediction and reconstruction errors exceed the dynamically set threshold, allowing for more adaptive and robust anomaly detection in varying maritime conditions.

\section{Experiment}
\subsection{Dataset}
In this paper, we use the \textit{OMTAD dataset} \cite{masek2021open}, which was constructed from the Australian Maritime Safety Authority (AMSA), using anonymized AIS data. We specifically focused on the Western Australian waters due to its high vessel traffic density, as it represents one of the busiest maritime routes in the region. Figure 2 shows the visualization of vessel trajectories in this area. The dataset contains fundamental AIS features: longitude, latitude, Speed over Ground (SOG), and Course over Ground (COG), with data points sampled at one-hour intervals. Vessel journeys were segmented based on stoppage (zero speed) or data gaps ($>$3.5 hours), and only tracks at least 10 hours long were retained. Missing data was interpolated using linear interpolation, with manual verification ensuring data quality. The final dataset includes 19,124 vessel tracks, categorized by vessel type (cargo, tanker, passenger), year, and month. While the original dataset included fishing vessels as a fourth category, we excluded them from our analysis as their distinctive movement patterns significantly differ from the other vessel types and could potentially be misclassified as anomalies in our model.

\begin{figure}[t]
\centering
\includegraphics[width=0.95\columnwidth]{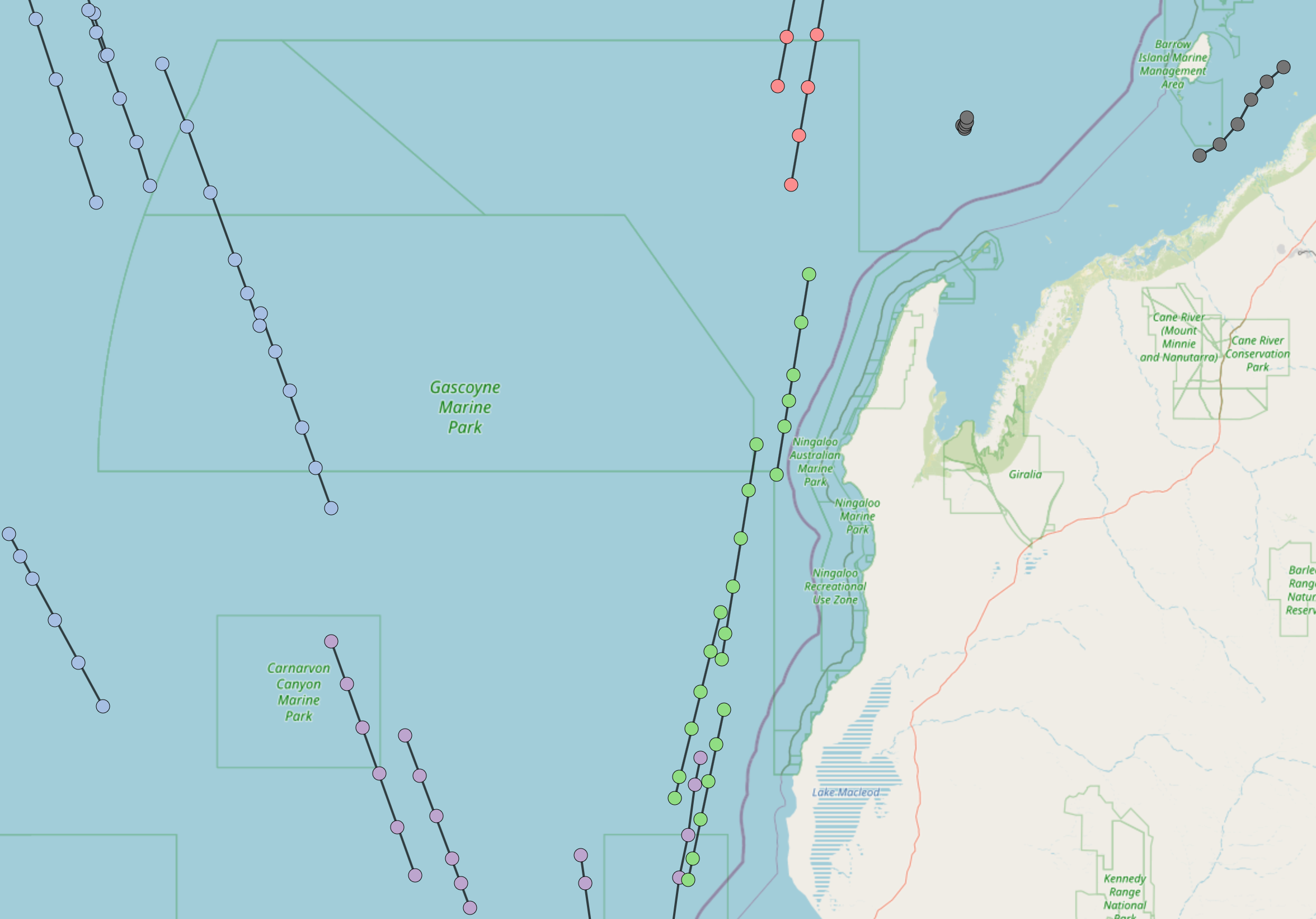}
\caption{ 
Visualization of maritime vessel trajectories from the OMTAD dataset in Western Australian waters, one of the most congested maritime areas in the region. The movement paths show vessel positions over a 5-hour period, with lines representing trajectories and points indicating vessel positions. The OPTICS algorithm identified distinct vessel clusters (shown in different colors), while gray trajectories represent non-clustered vessels classified as noise.
}
\label{fig2}
\end{figure}

\paragraph{Synthetic Anomaly Generation}
A major challenge in maritime anomaly detection is the lack of ground truth labels. In cases where labeled anomalies exist, they are typically annotated by domain experts, which is impractical for large-scale datasets. To address this issue, many existing approaches synthesize anomalies by injecting artificial perturbations into normal data. Inspired by this, we generate synthetic anomalies based on the \textit{Speed and Course Anomaly} (SCA) defined by \cite{liu2024ais}. SCA is characterized by abnormal changes in both speed and course, making it one of the most commonly observed anomalies in vessel trajectories.

To synthesize such anomalies, we model the statistical properties of normal vessel movements and introduce perturbations in the SOG and COG values. Specifically, we assume that the rate of change in SOG and COG follows a normal distribution:

\begin{equation}
    a \sim \mathcal{N}(\mu_a, \sigma_a^2), \quad \omega \sim \mathcal{N}(\mu_\omega, \sigma_\omega^2)
\end{equation}

where \( a_i = \frac{SOG_i - SOG_{i-1}}{\Delta t} \) and \( \omega_i = \frac{COG_i - COG_{i-1}}{\Delta t} \) denote the change rate of speed and course, respectively. Here, \( i \) represents the index of a data point in the trajectory, and \( \Delta t \) is the time interval between consecutive AIS messages.

To introduce synthetic anomalies, we perturb the speed and course change rates by sampling values that significantly deviate from their normal distribution:

\begin{equation}
    a^*_i = \mu_a + k \cdot \sigma_a, \quad \omega^*_i = \mu_\omega + k \cdot \sigma_\omega
\end{equation}

where \( k \) is a scaling factor that determines the severity of the anomaly. A typical choice is \( k > 3 \), ensuring that the new values fall outside the normal range (beyond the \( 99.7\% \) confidence interval).

Once the synthetic anomaly values \( a^*_i \) and \( \omega^*_i \) are generated, the corresponding speed and course values are updated iteratively:

\begin{equation}
    SOG^*_i = SOG_{i-1} + a^*_i \cdot \Delta t
\end{equation}

\begin{equation}
    COG^*_i = COG_{i-1} + \omega^*_i \cdot \Delta t
\end{equation}

where \( SOG^*_i \) and \( COG^*_i \) represent the modified speed and course values containing synthetic anomalies.

This process is applied to randomly selected trajectory points, ensuring that a subset of the dataset contains anomalous behavior. The generated anomalies effectively simulate real-world vessel irregularities, such as sudden acceleration, deceleration, or unexpected course deviations, making them valuable for training and evaluating anomaly detection models. 

\subsection{Experimental Setup}
All experiments were implemented using PyTorch 2.0 and PyTorch Geometric frameworks. Our architecture consists of a GCN encoder with 64 hidden channels and 32 latent dimensions, utilizing LayerNorm for stable training. For graph construction, we employed a sliding window approach with $h=10$ hours and 1-hour steps. Each node encodes a vessel state through five features: geographical coordinates, speed, and course decomposed into sine and cosine components. This decomposition of course angle was necessary to preserve the circular nature of angular data, as direct use of raw angles would create discontinuity at the 0/360-degree boundary.

The model was trained using the Adam optimizer with a learning rate of $0.001$ and weight decay of $1e$-$5$. We trained for a maximum of 100 epochs with early stopping (patience=$10$) to prevent overfitting. The loss function combines three components: forecasting loss (MSE), graph reconstruction loss, and KL divergence term. For synthetic anomaly generation, we set track\_anomaly\_ratio=$0.1$ ($10\%$ of tracks contain anomalies) and point\_anomaly\_ratio=$0.3$ ($30\%$ of points within anomalous tracks), with severity factor $k=3.5$.

The dataset was partitioned to ensure all synthetic anomalies appeared exclusively in the test set to maintain uncontaminated training data. Specifically, any trajectory containing $y=0$ values was allocated to the test set, while the remaining data was split into training ($90\%$) and validation ($10\%$) sets. Input features were normalized using per-graph standardization to ensure stable training.

\section{Conclusion and Future Work}

The proposed framework will be validated using AIS data, with comparative analyses conducted alongside existing anomaly detection models to evaluate performance. Furthermore, efforts will be made to refine the methodology through the use of adaptive graph construction techniques, with the aim of assessing its robustness and scalability across various maritime scenarios.

\bibliography{aaai25.bib}

\appendix
\section{Appendix}
\paragraph{Bounded Position Change Under Small Perturbations in SOG and COG} A key assumption in the anomaly generation process is that modifying SOG and COG does not significantly alter the vessel's position. The following theorem provides a theoretical justification for this assumption.

\noindent\textbf{Theorem 1.}  
For a sufficiently small perturbation \( \epsilon \), if the SOG and COG satisfy the following conditions:

\[
|a^*_i| \leq \epsilon SOG_{i-1}, \quad |\omega^*_i| \leq \epsilon COG_{i-1}
\]

then the resulting position change remains bounded by:

\[
|\Delta \phi^*| \leq \epsilon \Delta t, \quad |\Delta \lambda^*| \leq \epsilon \Delta t
\]

implying that the vessel remains effectively on the same trajectory.

\noindent
where
\begin{itemize}
    \item \( \Delta \phi^* \) and \( \Delta \lambda^* \) denote the changes in latitude and longitude, respectively.
\end{itemize}

Theorem 1 ensures that, under controlled perturbations, the trajectory remains unchanged, making it valid to generate anomalies by modifying only SOG and COG. Further proof is provided in the Appendix section.

\paragraph{Proof of Theorem 1} We start by defining the changes in latitude and longitude.  
Let \( R \) denote the Earth's radius. The latitude change \( \Delta \phi^* \) is computed using \( R \), while the longitude change \( \Delta \lambda^* \) is adjusted by \( R \cos\phi \) to account for the varying circumference of the Earth at different latitudes:

\[
\Delta \phi^* = \frac{SOG^*_i \cos(COG^*_i)}{R} \cdot \Delta t
\]

\[
\Delta \lambda^* = \frac{SOG^*_i \sin(COG^*_i)}{R \cos\phi} \cdot \Delta t
\]

Substituting the perturbed values of SOG and COG, we obtain:

\[
\Delta \phi^* = \frac{(SOG_{i-1} + a^*_i \cdot \Delta t) \cos(COG_{i-1} + \omega^*_i \cdot \Delta t)}{R} \cdot \Delta t
\]

\[
\Delta \lambda^* = \frac{(SOG_{i-1} + a^*_i \cdot \Delta t) \sin(COG_{i-1} + \omega^*_i \cdot \Delta t)}{R \cos\phi} \cdot \Delta t
\]

Using the Taylor series expansion for small \( x \), we approximate:

\[
\cos(COG^*_i) \approx \cos(COG_{i-1}) - \sin(COG_{i-1}) \cdot \omega^*_i \cdot \Delta t
\]

\[
\sin(COG^*_i) \approx \sin(COG_{i-1}) + \cos(COG_{i-1}) \cdot \omega^*_i \cdot \Delta t
\]

By substituting these approximations back into the expressions for \( \Delta \phi^* \) and \( \Delta \lambda^* \), we get:

\[
\Delta \phi^* \approx \frac{SOG_{i-1} \left( 1 + \frac{a^*_i}{SOG_{i-1}} \right) \cos COG_{i-1} }{R} \cdot \Delta t
\]

\[
\Delta \lambda^* \approx \frac{SOG_{i-1} \left( 1 + \frac{a^*_i}{SOG_{i-1}} \right) \sin COG_{i-1} }{R \cos \phi} \cdot \Delta t
\]

Since we assume:

\[
|a^*_i| \leq \epsilon SOG_{i-1}, \quad |\omega^*_i| \leq \epsilon COG_{i-1}
\]

it follows that:

\[
|\Delta \phi^*| \leq \epsilon \Delta t, \quad |\Delta \lambda^*| \leq \epsilon \Delta t
\]

which ensures that for a sufficiently small \( \epsilon \), the vessel remains effectively on the same trajectory.

\end{document}